\documentclass[11pt]{article}

\usepackage[preprint]{colm2026_conference}

\usepackage{latexsym}

\usepackage{microtype}
\usepackage{inconsolata}
\usepackage{graphicx}
\providecommand{\textasciislash}{/}

\usepackage{booktabs}
\usepackage{tabularx}
\usepackage{longtable}
\usepackage{hyperref}
\usepackage{xurl} %
\usepackage{enumitem}
\usepackage{xcolor}

\usepackage{fontspec}
\IfFontExistsTF{NotoSerifCJK-subset.otf}{
  \newfontfamily{\cjkfont}{NotoSerifCJK-subset.otf}
}{\IfFontExistsTF{HaranoAjiMincho-Regular.otf}{
  \newfontfamily{\cjkfont}{HaranoAjiMincho-Regular.otf}
}{\IfFontExistsTF{Noto Serif CJK KR}{
  \newfontfamily{\cjkfont}{Noto Serif CJK KR}
}{\IfFontExistsTF{Noto Serif KR}{
  \newfontfamily{\cjkfont}{Noto Serif KR}
}{
  \typeout{WARNING: no CJK font found.}
  \newcommand{\cjkfont}{}
}}}}

\IfFontExistsTF{NotoSerifDevanagari-subset.otf}{
  \newfontfamily{\devafont}{NotoSerifDevanagari-subset.otf}
}{\IfFontExistsTF{NewCM10Devanagari-Regular.otf}{
  \newfontfamily{\devafont}{NewCM10Devanagari-Regular.otf}
}{\IfFontExistsTF{Noto Serif Devanagari}{
  \newfontfamily{\devafont}{Noto Serif Devanagari}
}{\IfFontExistsTF{Noto Sans Devanagari}{
  \newfontfamily{\devafont}{Noto Sans Devanagari}
}{
  \typeout{WARNING: no Devanagari font found.}
  \newcommand{\devafont}{}
}}}}

\newcommand{\liltlogo}{\raisebox{0.1em}{\includegraphics[height=1.15em]{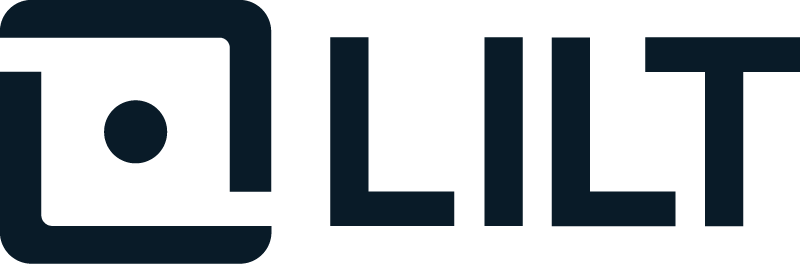}}}

\definecolor{todoyellow}{RGB}{255,243,176}

\usepackage{subcaption}   %
\usepackage{listings}     %
\usepackage{pifont}       %

\definecolor{qcpass}{HTML}{7DB343}
\definecolor{qcfail}{HTML}{F34135}
\newcommand{\passmark}{\textcolor{qcpass}{\ding{51}}}
\newcommand{\failmark}{\textcolor{qcfail}{\ding{55}}}

\lstdefinestyle{instr}{
  basicstyle=\ttfamily\tiny,
  breaklines=true, breakindent=0pt, breakautoindent=false,
  columns=fullflexible, keepspaces=true, upquote=true,
  frame=single, framesep=4pt, rulecolor=\color{gray!60},
  xleftmargin=3pt, xrightmargin=3pt, aboveskip=2pt, belowskip=2pt,
  extendedchars=true,
  literate=%
    {č}{{\v c}}1 {Č}{{\v C}}1 {ě}{{\v e}}1 {Ě}{{\v E}}1
    {š}{{\v s}}1 {Š}{{\v S}}1 {ž}{{\v z}}1 {Ž}{{\v Z}}1
    {ř}{{\v r}}1 {Ř}{{\v R}}1 {ň}{{\v n}}1 {ď}{{\v d}}1 {ť}{{\v t}}1
    {á}{{\'a}}1 {í}{{\'\i}}1 {é}{{\'e}}1 {ú}{{\'u}}1 {ů}{{\r u}}1
    {ý}{{\'y}}1 {ó}{{\'o}}1 {ü}{{\"u}}1 {ö}{{\"o}}1 {ä}{{\"a}}1
    {ß}{{\ss}}1 {§}{{\S}}1 {—}{{---}}1 {–}{{--}}1
}

\title{Terminal-Bench-LILT: Multilingual Agentic Coding Benchmark Grounded in Language, Region, and Culture}

\author{%
{\bf Yunsu Kim \quad Kaden Uhlig \quad Ashwin Purohit \quad Milind Agarwal \quad Patrick Simianer} \\
{\bf Anil Arslan \quad Kiarash Mokhtari \quad Thomas Zenkel \quad Johannes Mosig} \\
{\bf Gabriel Bretschner \quad Shamik Bose \quad Joern Wuebker \quad John DeNero} \\[4pt]
LILT, Inc. \\
\texttt{contact@lilt.com}
}

\begin{document}

\maketitle

\lhead{\liltlogo}

\begin{abstract}
Most evaluations for coding agents are conducted exclusively in English, which does not reflect real-world multilingual deployment.
We present Terminal-Bench-LILT, a suite of 300 authentic coding tasks in 10 languages: Arabic, Czech, German, Spanish, Hindi, Japanese, Korean, Serbian, Turkish, and Chinese.
Each task targets issues specific to non-English software development that have no direct English equivalent, e.g., internationalization, encoding, text normalization, and cultural conventions.
All tasks are authored by native-speaker programmers and validated through a multi-stage quality control pipeline.
Evaluation of six frontier models reveals that even the strongest model reaches only 63.1\% pass rate, with many tasks unsolved by any model.
Performance varies substantially by language and does not track general coding benchmark rankings, highlighting that multilingual coding competence is a distinct and underexplored capability axis.
Sample tasks are available at \url{https://github.com/lilt/terminal-bench-lilt}.
\end{abstract}

\section{Introduction}
\label{sec:intro}

Software development is a global field in which English serves as the common working language: open-source projects are coordinated in English \citep{kocetkov2023stack,lozhkov2024starcoder2,bhuiyan2026write}, and community resources such as Stack Overflow are written and answered largely in English \citep{stackoverflow_nonenglish}.
Most programmers therefore read, write, and discuss code in English, and the models and tools built on this ecosystem are trained and tuned primarily for English and perform best in it \citep{githubcopilot_agents}.
The benchmarks used to evaluate these agents follow the same pattern: beyond being written almost entirely in English, they implicitly encode the data, locales, and conventions of an English-speaking development context \citep{chen2021codex,jimenez2024swebench,jain2025livecodebench,merrill2026terminalbench}.

However, many developers are not native English speakers \citep{slashdata2025population,github2025octoverse} and solve problems outside an English-speaking context.
Particularly in local markets, developers work with native-language data, configure locale-specific settings, and build services that fit local conventions and culture.
These are real engineering challenges that rarely surface in English-speaking contexts, and they directly shape how non-English-speaking developers experience coding agents.
The ideal coding agent proactively identifies and resolves implicit technical and cultural assumptions.
This is taken for granted in English-centric tasks, while in non-English contexts agents need explicit guidance to avoid pitfalls native speakers consider obvious \citep{shen-etal-2024-understanding,naous-etal-2024-beer,myung2024blend}.
This raises the expertise required of users and makes the agent tedious to use.

No existing coding benchmark evaluates agents on this dimension, so we built Terminal-Bench-LILT, a suite of 300 terminal-based coding tasks in 10 languages across three difficulty tiers.
Native programmers authored each task from a problem they had actually encountered, then put it through layered quality control combining automated checks and human review.

\begin{figure*}[t]
\centering
\begin{subfigure}[t]{0.46\textwidth}
\vspace{0pt}%
\begin{lstlisting}[style=instr]
I'm on the Tokyo NOC (network operations center) team, and I need to fit our daily-status log lines cleanly into the fixed 40-column monitoring terminal we print them to.

...

- Strip ANSI escape sequences. They take 0 display columns and must not appear in the output.
- Convert halfwidth katakana to its fullwidth equivalent. The halfwidth voiced mark and semi-voiced mark merge into the preceding kana. The prolonged sound mark becomes fullwidth.

...
\end{lstlisting}
    \caption{Instruction}
    \label{fig:ja-instruction}
  \end{subfigure}\hfill
  \begin{subfigure}[t]{0.51\textwidth}
    \vspace{0pt}%
    \centering
    \includegraphics[width=\textwidth]{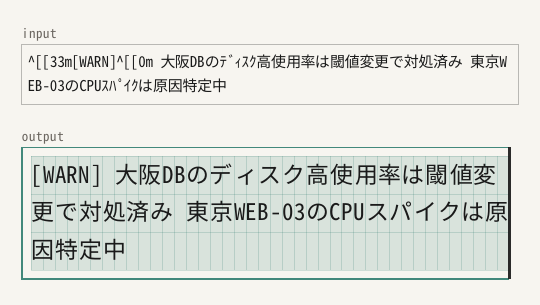}
    \caption{Example input/output.}
    \label{fig:ja-wrapping}
  \end{subfigure}
\caption{The \texttt{ja-terminal-width} task from Terminal-Bench-LILT. The instruction in (a) is abridged for clarity. In (b), each grid cell is one display column of the terminal.}
\label{fig:ja-terminal-width}
\end{figure*}

Figure \ref{fig:ja-terminal-width} shows a task from Terminal-Bench-LILT.
The agent is asked to fit Japanese log lines into a fixed 40-column terminal.
Each line mixes Japanese characters, ASCII, emoji, and terminal color escapes, which occupy different numbers of columns.
Further edge cases arise naturally from Japanese text itself, for example:
\begin{enumerate}\itemsep0em
  \item Unicode variation selectors, often used in names to force a specific visual variant, are not displayed and must be left out of the width calculation.
  Counting code points, as Python's \texttt{len()} does, gives each one a column.
  \item Kanji have compatibility ideographs that look identical but have different bytes, e.g., \mbox{{\cjkfont 神} = U+FA19} and \mbox{{\cjkfont 神} = U+795E}.
  Applying Unicode NFKC without care merges them into the more common form, altering text the task only asked to reformat.
\end{enumerate}
Such issues are routine for a developer who handles Japanese terminal output and tedious to spell out.
The instruction therefore asks for it only implicitly, while the tests check exactly these points.
The task thus precisely measures a smooth user experience for a native developer working in Japanese.

The rest of the paper is organized as follows.
Section \ref{sec:related} reviews prior coding benchmarks and related work.
Section \ref{sec:tb-lilt} describes the design, construction, and quality control of Terminal-Bench-LILT.
Section \ref{sec:exp} evaluates frontier models on the benchmark and analyzes the results.
Section \ref{sec:conc} concludes, followed by an appendix listing sample tasks.

\section{Related Work}
\label{sec:related}

\paragraph{Agentic Coding Benchmarks}
Evaluation of code generation has moved from single-function synthesis \citep{chen2021codex} toward repository-level and agentic settings: SWE-bench \citep{jimenez2024swebench} poses real GitHub issues, LiveCodeBench \citep{jain2025livecodebench} provides contamination-free competitive problems, and Terminal-Bench \citep{merrill2026terminalbench} evaluates agents on command-line tasks.
These benchmarks are all English-only; we build on Terminal-Bench and extend its task format to ten non-English languages.

\paragraph{Multilingual Coding Benchmarks}
Many coding benchmarks that call themselves \emph{multilingual} vary the \emph{programming} language of the code while keeping the instruction in English \citep{cassano2023multiple,athiwaratkun2023mbxp,zan2025multiswebenchmultilingualbenchmarkissue}.
Others do vary the human language, but only in the instruction, translating English coding tasks into other natural languages \citep{raihan2025mhumaneval,peng2024humanevalxl,wang2023mconala,wang2024exploringmultilingualbiaslarge,maps2025}.
They primarily measure model comprehension of translated prompts, not how useful a model is for developers in a non-English context.

In coding, this comprehension axis seems to be an inconsistent driver of difficulty.
For instance, \citet{maps2025} translate SWE-bench \citep{jimenez2024swebench} into ten languages, but the mean non-English performance shows only a $1.3\%$ relative drop from English.
Meanwhile, \citet{wang2024exploringmultilingualbiaslarge} report a relative decrease of at least $13\%$ in pass rate on HumanEval-X \citep{zheng2023codegeex} tasks translated into Chinese.
In these cases, it is hard to distinguish between errors introduced by translation and genuine limits of model capability; moreover, all tasks are drawn from an English context.

Our benchmark is orthogonal to both lines of work: the programming language is incidental, and we vary the human language together with its regional and cultural context.
Our tasks are authored directly in a given language by native speakers, avoiding translation artifacts, and we keep only tasks that are genuinely language- and culture-specific, enabling a targeted evaluation of models' usefulness for non-English developers.

\paragraph{Reasoning Language}
One hypothesis is that a translated instruction leaves the underlying problem unchanged: a model can read a non-English prompt yet still reason in English \citep{schut2025multilingual}.
\citet{barua2026longchainofthoughtreasoninglanguages} find that reasoning in English outperforms reasoning in the target language, with the gap widening on multi-step tasks.
For code generation, \citet{nishigata-etal-2025-non} exploit this directly, tuning models to attach an English chain-of-thought to non-English instructions.
Difficulty from the instruction language is thus limited, and such tuning can mitigate it further.
Our tasks instead rest on the locale-specific knowledge the instructions leave implicit --- a gap that reasoning alone does not close (Section~\ref{sec:reasoning}).

\paragraph{Cultural and Locale Knowledge}
Benchmarks of everyday cultural knowledge expose large disparities between well- and under-represented cultures: \citet{myung2024blend} report a best-to-worst culture gap of up to 57~percentage points for GPT-4 on short-answer questions, and related work documents cultural bias \citep{naous-etal-2024-beer} and the limits of cultural commonsense \citep{shen-etal-2024-understanding} in LLMs.
These benchmarks, however, probe such knowledge in plain question answering, not in the work that models are actually used for.
Our benchmark tests the same knowledge in agentic coding, where an agent passes only if its code handles the local convention correctly.

\section{Terminal-Bench-LILT}
\label{sec:tb-lilt}

Terminal-Bench-LILT has 300 tasks spanning 10~languages, with 30 tasks in each language.
Each task runs under Harbor \citep{Harbor_Framework} in an isolated environment with deterministic verification.
It follows the original Terminal-Bench task structure \citep{merrill2026terminalbench}, with additional fields and documentation for multilingual challenges:

\begin{itemize}\itemsep0em
  \item \textbf{Instruction} describes a real-world coding problem requiring the agent to write a solution script. It is provided in both the native language and English.
  \item \textbf{Environment} (Docker configuration) together with any native-language data assets required by the task.
  \item \textbf{Verifier} with deterministic tests.
  \item \textbf{Oracle solution} accepted by the verifier.
  \item \textbf{Metadata} containing brief descriptions of the task, solution, and verifier, together with the ISO~639-1 language code, challenge category and subcategory, difficulty, and infrastructure configuration.
  \item \textbf{README} explaining the motivation behind the task design, its realism, and its linguistic and technical challenges in detail.
\end{itemize}

Instructions were written \emph{from scratch in the native target language} first.
Many tasks are grounded in locale-specific data, such as files in legacy encodings and regional date formats, and are therefore most naturally described in the language in which the data was produced.
This approach also reflects how users instruct coding agents in their native languages, using colloquial phrasing and domain-specific vocabulary.
Writing the instructions in English first and then translating them, as is common in prior work (Section~\ref{sec:related}), could introduce translation artifacts and make the resulting tasks less representative of real-world use.

An English translation of the native instruction is also provided, making it accessible to non-native speakers and enabling controlled experiments that isolate the effect of instruction language from task difficulty (Section \ref{sec:ablation}).

\subsection{Task Design}

\begin{table*}[!t]
  \centering
  \small
  \begin{tabularx}{\textwidth}{>{\raggedright\arraybackslash}p{80pt} >{\raggedright\arraybackslash}p{109pt} X}
  \toprule
  \textbf{Category} & \textbf{Subcategory} & \textbf{Scope / Examples} \\
  \midrule
  Internationalization & Locale-based Templating & Plurals, numbers, suffixes, date formats, Korean \emph{man} numbering \\
                       & Character Rendering & RTL, Arabic shaping, Simplified vs.\ Traditional Chinese, Cyrillic \\
  \midrule
  Interaction with Environment & Encodings & Non-Unicode/legacy file encodings, conversion \\
                       & System Configuration & CJK input method debugging, RTL shell tools, locale-correct sorting \\
  \midrule
  Text Conversion      & Normalization & NFC/NFKC normalization, mojibake correction \\
                       & ML \& Data Processing & Case augmentation, tokenizer training, typo simulation \\
                       & Variant Conversion & Script conversion, German spelling reform \\
  \midrule
  Text Handling        & Document Search & Stemming, stop word removal for agglutinative languages \\
                       & Unicode Confusion Lists & Anti-spam character confusion in Cyrillic-first applications \\
                       & User Authentication & Normalized username/password comparison, DB migration \\
  \midrule
  Cultural \& Other    & Calendar/Date Conversion & Lunar, Hijri, Umm al-Qura, holiday calculation \\
                       & Other & Culture/Region-specific issues that cannot be categorized into existing subcategories \\
  \bottomrule
  \end{tabularx}
  \caption{Challenge taxonomy for Terminal-Bench-LILT tasks.}
  \label{tab:taxonomy}
\end{table*}

\begin{table*}[!t]
  \centering
  {\small % Auto-generated by data/generate_taxonomy_stats.py. Do not edit by hand.
\begin{tabular}{lccccccccccc}
\toprule
\textbf{Category} & \textsc{ar} & \textsc{cs} & \textsc{de} & \textsc{es} & \textsc{hi} & \textsc{ja} & \textsc{ko} & \textsc{sr} & \textsc{tr} & \textsc{zh} & \textbf{Total} \\
\midrule
Internationalization & 10 & 7 & 3 & 2 & 7 & 5 & 6 & 6 & 5 & 1 & 52 \\
Interaction with Environment & 2 & 2 & 5 & 3 & 3 & 3 & 1 & 1 & 2 & 3 & 25 \\
Text Conversion & 10 & 3 & 3 & 11 & 6 & 7 & 2 & 15 & 12 & 7 & 76 \\
Text Handling & 1 & 1 & 2 & 2 & 4 & 2 & 1 & 3 & 2 & 0 & 18 \\
Cultural \& Other & 7 & 17 & 17 & 12 & 10 & 13 & 20 & 5 & 9 & 19 & 129 \\
\midrule
\textbf{Total} & 30 & 30 & 30 & 30 & 30 & 30 & 30 & 30 & 30 & 30 & \textbf{300} \\
\bottomrule
\end{tabular}
}
  \caption{Distribution of tasks by challenge category and language.}
  \label{tab:taxonomy-stats}
\end{table*}

Each task includes one or more challenges specific to its target language, region, or culture.
We assign every task a primary category and subcategory from the challenge taxonomy in Table~\ref{tab:taxonomy}; Table~\ref{tab:taxonomy-stats} shows the distribution of tasks by category and language.
Descriptions of sample tasks are given in Appendix \ref{sec:appendix-task-list}.

Regardless of category, every task follows three design principles:
\begin{enumerate}\itemsep0em
  \item A task should reflect a problem that a programmer in the target locale could plausibly encounter.
  Its data, formats, and workflows should match those \emph{actually used in practice}.
  \item A task’s difficulty should arise from \emph{locale-specific elements}, not from general engineering complexity.
  A task that merely restates an ordinary programming problem in another language does not qualify.
  \item Locale-specific challenges should be \emph{embedded implicitly} in the engineering problem rather than stated directly.
  This tests whether agents can supply the local knowledge that native developers take for granted, without requiring users to spell it out.
\end{enumerate}

The third principle is particularly important for cultural challenges where the task involves everyday knowledge in the target region beyond technical conventions such as encoding or sorting order.
When the required knowledge is too specialized, e.g., a traffic rule specific to one state, the task becomes asking whether a model happens to know it, even though modern agents can retrieve such facts through web search or reference files.
We avoid such esoteric knowledge recall and focus on native user experience by keeping the two rules below:

\begin{itemize}\itemsep0em
  \item Any challenge should be \emph{widely known} to ordinary adults or programmers in the target country.
  \item When an expert domain such as law or medicine is involved, it should serve only as realistic \emph{context}, never as the challenge itself.
  This is realized by citing or attaching the exact reference in the instruction.
\end{itemize}

\begin{figure*}[t]
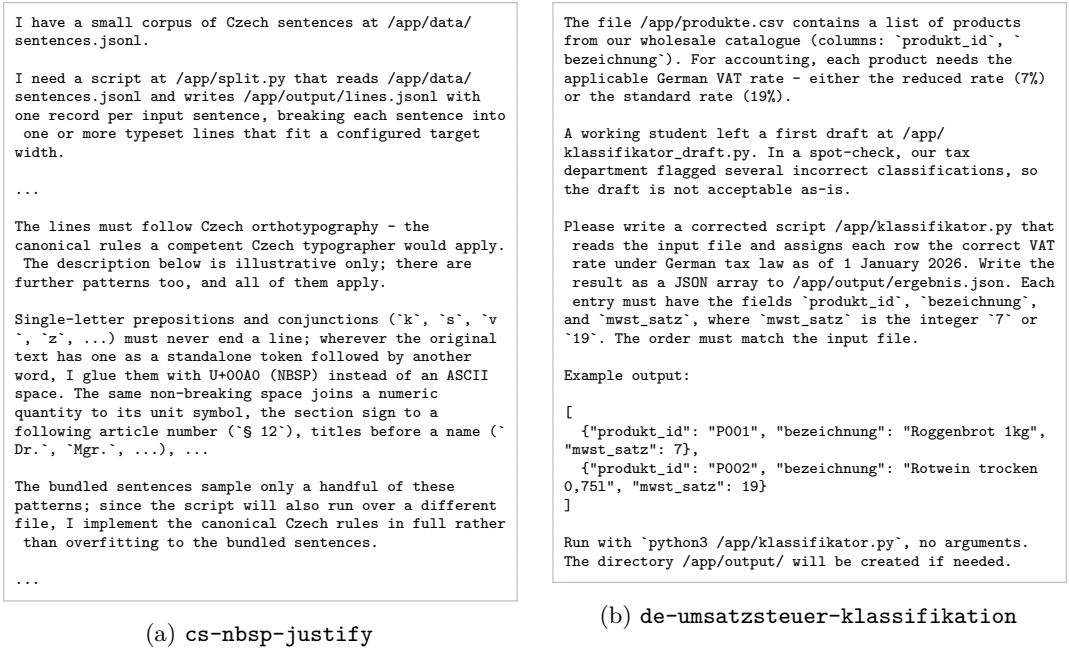

  \centering
  \begin{subfigure}[t]{0.48\textwidth}
\begin{lstlisting}[style=instr]
I have a small corpus of Czech sentences at /app/data/sentences.jsonl.

I need a script at /app/split.py that reads /app/data/sentences.jsonl and writes /app/output/lines.jsonl with one record per input sentence, breaking each sentence into one or more typeset lines that fit a configured target width.

...

The lines must follow Czech orthotypography - the canonical rules a competent Czech typographer would apply. The description below is illustrative only; there are further patterns too, and all of them apply.

Single-letter prepositions and conjunctions (`k`, `s`, `v`, `z`, ...) must never end a line; wherever the original text has one as a standalone token followed by another word, I glue them with U+00A0 (NBSP) instead of an ASCII space. The same non-breaking space joins a numeric quantity to its unit symbol, the section sign to a following article number (`§ 12`), titles before a name (`Dr.`, `Mgr.`, ...), ...

The bundled sentences sample only a handful of these patterns; since the script will also run over a different file, I implement the canonical Czech rules in full rather than overfitting to the bundled sentences.

...
\end{lstlisting}
    \caption{\texttt{cs-nbsp-justify}}
    \label{fig:boundary-case1}
  \end{subfigure}\hfill
  \begin{subfigure}[t]{0.48\textwidth}
\begin{lstlisting}[style=instr]
The file /app/produkte.csv contains a list of products from our wholesale catalogue (columns: `produkt_id`, `bezeichnung`). For accounting, each product needs the applicable German VAT rate — either the reduced rate (7%) or the standard rate (19%).

A working student left a first draft at /app/klassifikator_draft.py. In a spot-check, our tax department flagged several incorrect classifications, so the draft is not acceptable as-is.

Please write a corrected script /app/klassifikator.py that reads the input file and assigns each row the correct VAT rate under German tax law as of 1 January 2026. Write the result as a JSON array to /app/output/ergebnis.json. Each entry must have the fields `produkt_id`, `bezeichnung`, and `mwst_satz`, where `mwst_satz` is the integer `7` or `19`. The order must match the input file.

Example output:

[
  {"produkt_id": "P001", "bezeichnung": "Roggenbrot 1kg", "mwst_satz": 7},
  {"produkt_id": "P002", "bezeichnung": "Rotwein trocken 0,75l", "mwst_satz": 19}
]

Run with `python3 /app/klassifikator.py`, no arguments. The directory /app/output/ will be created if needed.
\end{lstlisting}
    \caption{\texttt{de-umsatzsteuer-klassifikation}}
    \label{fig:boundary-case2}
  \end{subfigure}
  \caption{Instructions for two sample tasks with different challenge scopes: (a) a bounded challenge and (b) an unbounded one fenced to a finite range.}
  \label{fig:boundary-examples}
\end{figure*}

\subsection{Test Design}

Each task comes with a test suite combining basic integrity checks (e.g., whether a solution script was created), sample probes using the files provided with the task, and unseen probes using new inputs.
The unseen probes test generalization and discourage hard-coded solutions, but remain within the intended challenge.

Every tested behavior must be \emph{inferable} from either the instruction, the input data, or common knowledge in the target locale.
The instruction should make clear that the provided files are examples rather than the complete input set; otherwise, unseen probes would test behavior that cannot be inferred from the instruction.
Requirements unrelated to the challenge, such as the output format, must be stated unambiguously.

The scope of each challenge must also be testable.
When the challenge covers a \emph{bounded} set of cases, the instruction may show only some of them and withhold the rest for testing, requiring a general solution.
For example, in Figure \ref{fig:boundary-case1}, the Czech non-breaking-space rules form a finite, closed set, so the instruction gives only a few example patterns and a small sample of sentences; the held-out sentences exercise the patterns the samples never show, catching solutions that overfit to the visible ones.
When the cases are \emph{unbounded}, the instruction and tests must instead define a finite range; otherwise the challenge is excluded.
For example, in Figure \ref{fig:boundary-case2}, the German VAT rates are unbounded legislated carve-outs that no general solution captures, so the instruction fixes a finite range (a specific legal snapshot and product list) and the tests check only that range.

\subsection{Quality Control}
\label{sec:qc}

\begin{figure*}[t]
\centering
\includegraphics[width=\textwidth]{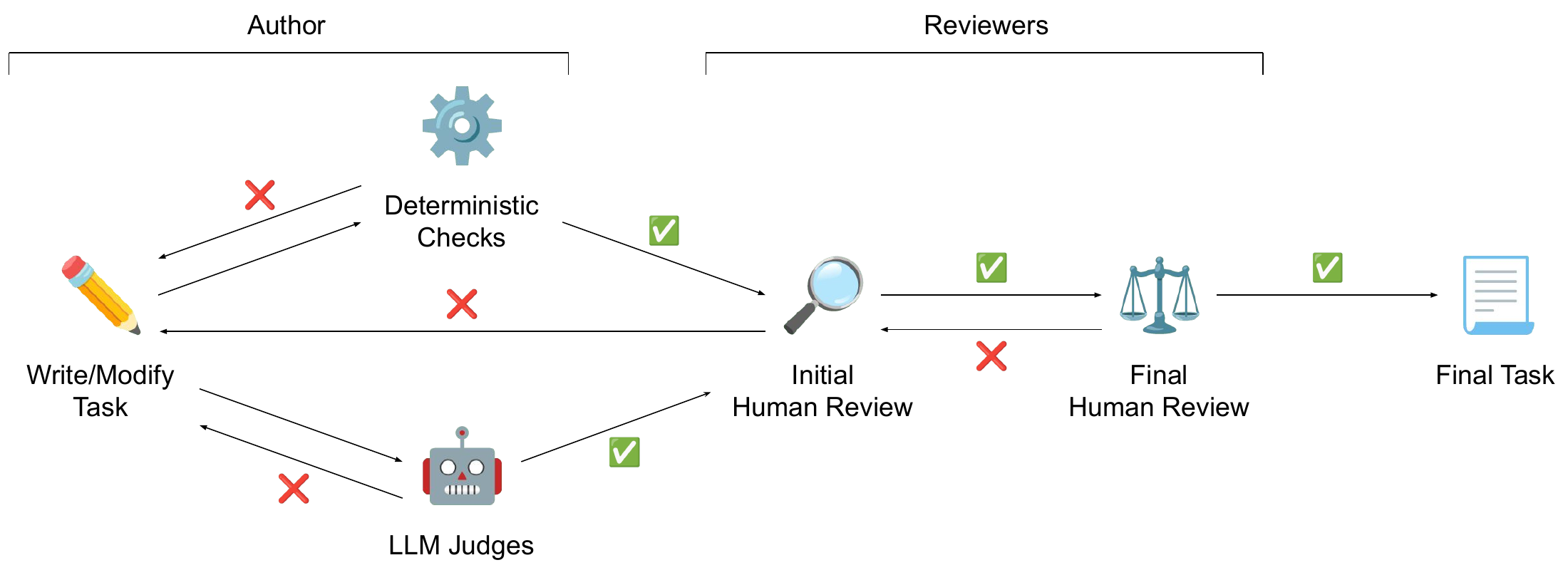}
\caption{Quality control workflow for task submissions. \passmark{} and \failmark{} indicate whether a task passes or fails each stage, respectively. The responsible party is shown above each stage.}
\label{fig:qc-workflow}
\end{figure*}

Every task passes four review layers before entering the benchmark.
The layers are ordered by cost: automated checks run on every commit by the task author, and human reviewers see only the tasks that survive them.
Figure~\ref{fig:qc-workflow} shows the full workflow for quality control.

\begin{enumerate}\itemsep0em
  \item \textbf{Deterministic checks} validate the task's structure and execution setup, including its file layout, Docker environment, metadata fields, and whether the oracle solution passes all tests.
  \item \textbf{LLM judges} use Gemini~3 Flash to assess the task's semantic properties, such as parity between the native-language and English instructions, alignment with the claimed task category, and whether agents can infer each tested behavior from the task materials or common local knowledge.
  \item \textbf{Initial human review} re-audits the automated checks and assesses criteria that are difficult to automate: e.g., compliance with the AI-use policy, realism of the coding scenario, and clarity of constraints.
  \item \textbf{Final human review} independently repeats the initial review, checks for overlap with existing tasks, and curates the final set for balanced topic and difficulty distributions, recommending difficulty adjustments when needed.
\end{enumerate}

At any stage, if we found that a task could not be made valid and realistic with reasonable effort, we rejected it.
Overall, about 25\% of reviewed tasks were rejected.

\paragraph{Review Agents}
Reviewing a task at this level of detail requires substantial time and sustained attention.
We therefore built two review agents to assist human reviewers: one assesses each task on its own, while the other compares it with existing tasks in the benchmark.
Each delegates subtasks to specialized agents and combines their findings into a single report with concrete citations.
We equipped the agents with deterministic tools whenever possible, e.g., to prepare execution logs and failure summaries, so the LLMs can focus on judgment.
Since deployment, we have periodically used human reviewer feedback on agent outputs to refine the rubrics, aligning automated review with human judgment and reducing review time.

\begin{figure*}[t]
\centering
\includegraphics[width=\textwidth]{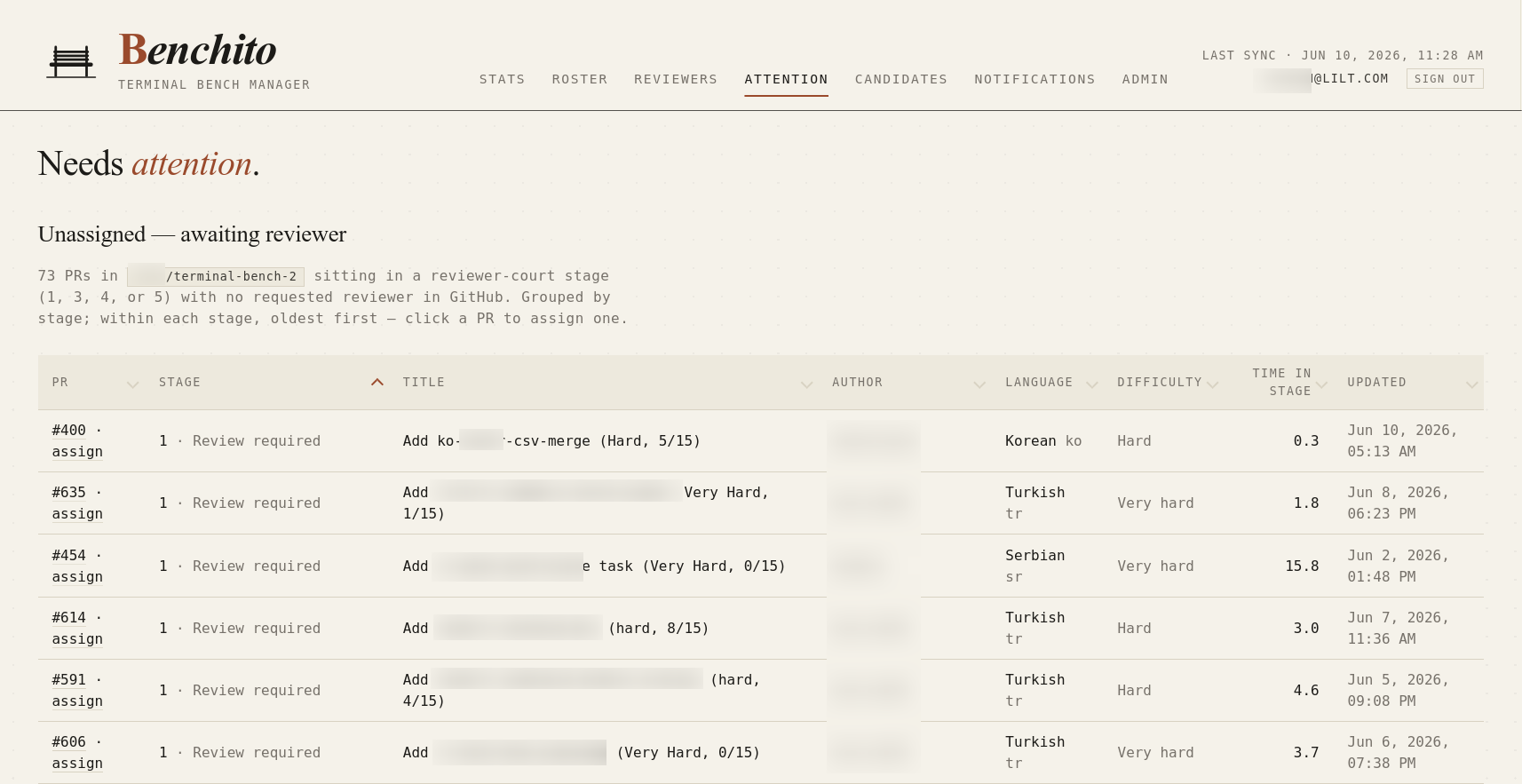}
\caption{Workflow management tool for Terminal-Bench-LILT.}
\label{fig:benchito}
\end{figure*}

\paragraph{Tooling}
To streamline the multi-stage review process for hundreds of tasks, we developed an internal management application \emph{Benchito} which tracks the stage of each submitted task (Figure~\ref{fig:benchito}).
Program managers can filter and sort tasks by language, difficulty, author, or reviewer and compare task counts for a selected period with the target distribution.
It is integrated with the project's communication channels and notifies the responsible author or reviewer when a task changes state or fails an automated check.
Access is restricted to the organization, with an audit trail of every action for accountability.

\subsection{Team Management}

Task quality depends on the expertise of both authors and reviewers.
We therefore used explicit selection criteria and structured training.

\paragraph{Author Selection}
For each language, we screened native-speaker candidates for backgrounds in computer science or engineering and for multilingual experience.
Candidates who met these criteria completed an assessment of their ability to design tasks that challenge LLMs, which our reviewers evaluated manually.
Fewer than 10\% of these candidates passed.

\paragraph{Reviewer Selection}
Initial reviewers were either research scientists with backgrounds in machine learning and NLP or task authors who had developed more than ten Terminal-Bench-LILT tasks.
Final reviewers were senior research scientists and engineers with extensive LLM benchmarking experience.

\paragraph{Onboarding and Training}
New task authors completed a first-day checklist covering access provisioning and tool setup.
Each then attended a one-hour training session on task architecture, Harbor, the taxonomy, example tasks, quality criteria, and the task submission workflow.
Before authoring tasks, they were required to demonstrate their understanding of the training materials, and reviewers provided detailed feedback on their first submission.
Reviewers remained available through a real-time channel for ongoing feedback and held targeted calibration sessions when needed.

\subsection{Semi-Automatic Task Creation}
\label{sec:auto-creation}
For experimental purposes and to supplement our dataset, we created a limited number of tasks through a partially automated workflow.
We built an agent equipped with specialized skills that operates in two stages:

\begin{enumerate}\itemsep0em
  \item \textbf{Idea generation.} First, the agent researches linguistic and cultural software conventions from the web, extracts candidate challenges, and removes ideas similar to accepted or rejected tasks.
  The remaining ideas are ranked by general quality and challenge specificity; a human reviewer approves the shortlist before development begins.
  This stage runs once per language.
  \item \textbf{Task generation.} For each approved idea, the agent produces a progress file with a development checklist.
  Following this checklist iteratively, it implements the solution and unit tests, runs automated checks, and calibrates difficulty from agent trajectories.
\end{enumerate}
After calibration, each created task undergoes the same automated checks and review process as tasks written by humans (Section~\ref{sec:qc}), with an additional initial stage in which a native speaker of the target language reviews only its linguistic quality.
This process produced approximately 10\% of the Terminal-Bench-LILT tasks.

\section{Experiments}
\label{sec:exp}

On the full Terminal-Bench-LILT set, we evaluated six frontier LLMs across three providers: Claude Opus~4.7 \citep{claude47_2026} and 4.8 \citep{claude48_2026} from Anthropic, GPT-5.4 \citep{gpt54_2026} and 5.5 \citep{gpt55_2026} from OpenAI, and Gemini~3.1~Pro \citep{gemini31_2026} and 3.5~Flash \citep{gemini35_2026} from Google.
Thinking/Reasoning level was set to the provider's default for each model, following the same convention as \citet{merrill2026terminalbench}.
The agent was fixed to the \texttt{terminus-2} harness and each task-model pair was run five times.
We also manually inspected the failing trials to distinguish true language-specific failures from infrastructure issues, e.g., random crashes, OOMs, API errors, or harness failures; any trial affected by the latter was re-run so that all reported results reflect genuine task attempts.

\begin{figure}[!t]
  \centering
  \includegraphics[width=\columnwidth]{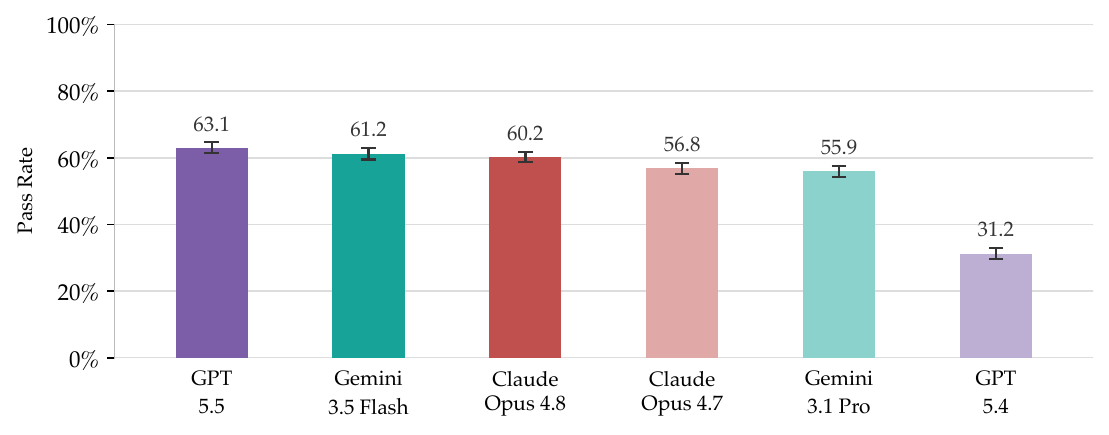}
  \caption{Task pass rate per model on Terminal-Bench-LILT, sorted by rate.
  Error bars show 95\% confidence intervals.}
  \label{fig:resolution-rate}
\end{figure}

\subsection{Main Results}
\label{sec:main}

Figure~\ref{fig:resolution-rate} shows the aggregate pass rate across all trials for each model.
GPT-5.5, Gemini~3.5~Flash, and Claude Opus~4.8 form the leading group, while Claude Opus~4.7 and Gemini~3.1~Pro follow, with GPT-5.4 far behind.
The newer OpenAI model improves dramatically over its predecessor (a 31.9-point gain), while Google's and Anthropic's newer models improve more modestly, by 5.3 and 3.4 points.
Even the strongest model fails 36.9\% of trials, indicating that the benchmark is not saturated and leaves substantial room for improvement.

\paragraph{Difficulty Calibration}
We used the same trial results to estimate task difficulty following \citet{merrill2026terminalbench}.
For each task, we pooled its 30 trial outcomes across the six models and computed the overall pass rate.
We classified tasks with pass rates of at least 66.7\% as easy, rates from 33.3\% to below 66.7\% as medium, and rates below 33.3\% as hard.
Table~\ref{tab:dataset-stats} shows an approximate 4:3:3 ratio of easy, medium, and hard tasks.

\begin{table*}[!t]
  \centering
  {\small
  \begin{tabular}{lccccccccccc}
  \toprule
  \textbf{Difficulty} & \textsc{ar} & \textsc{cs} & \textsc{de} & \textsc{es} & \textsc{hi} & \textsc{ja} & \textsc{ko} & \textsc{sr} & \textsc{tr} & \textsc{zh} & \textbf{Total} \\
  \midrule
  % inlined from data/colm_difficulty_stats.tex by build_arxiv.py
Easy & 16 & 17 & 13 & 12 & 8 & 13 & 12 & 16 & 17 & 9 & 133 \\
Medium & 9 & 4 & 6 & 12 & 12 & 8 & 9 & 7 & 6 & 13 & 86 \\
Hard & 5 & 9 & 11 & 6 & 10 & 9 & 9 & 7 & 7 & 8 & 81 \\
\midrule
\textbf{Total} & 30 & 30 & 30 & 30 & 30 & 30 & 30 & 30 & 30 & 30 & \textbf{300} \\
  \bottomrule
  \end{tabular}}
  \caption{Distribution of tasks by language and empirical difficulty.}
  \label{tab:dataset-stats}
  \end{table*}

\subsection{Performance Breakdown}

\begin{table*}[!t]
  \centering
  {\small\setlength{\tabcolsep}{5pt}% Auto-generated by data/generate_results.py. Do not edit by hand.
\begin{tabular}{l ccccccccccc}
\toprule
\textbf{Model} & \textsc{ar} & \textsc{cs} & \textsc{de} & \textsc{es} & \textsc{hi} & \textsc{ja} & \textsc{ko} & \textsc{sr} & \textsc{tr} & \textsc{zh} & $\sigma_{\mathrm{lang}}$ \\
\midrule
GPT-5.5 & 69.3 & 60.7 & 51.3 & 74.0 & 51.3 & 70.7 & 58.0 & 72.7 & 69.3 & 53.3 & 8.7 \\
Gemini 3.5 Flash & 62.7 & 64.0 & 56.7 & 60.7 & 52.0 & 57.3 & 58.7 & 66.0 & 69.3 & 64.7 & 4.9 \\
Claude Opus 4.8 & 59.3 & 68.7 & 50.0 & 68.7 & 52.7 & 52.7 & 57.3 & 61.3 & 66.7 & 64.7 & 6.6 \\
\midrule
$\sigma_{\mathrm{model}}$ & 4.2 & 3.3 & 2.9 & 5.5 & 0.5 & 7.6 & 0.5 & 4.7 & 1.3 & 5.3 & -- \\
\bottomrule
\end{tabular}
}
  \caption{Per-language pass rates. $\sigma_{\mathrm{lang}}$ and $\sigma_{\mathrm{model}}$ are the standard deviations (in percentage points) across languages and across models, respectively.}
  \label{tab:competitor-langs}
  \end{table*}

\begin{table*}[!t]
  \centering
  {\small % Auto-generated by data/generate_category_passrates.py. Do not edit by hand.
\setlength{\tabcolsep}{0pt}
\begin{tabularx}{\textwidth}{l @{\hspace{6pt}} >{\centering\arraybackslash}X >{\centering\arraybackslash}X >{\centering\arraybackslash}X >{\centering\arraybackslash}X >{\centering\arraybackslash}X}
\toprule
 & International- & Interaction w/ & Text & Text & Cultural \& \\
\textbf{Model} & ization & Environment & Conversion & Handling & Other \\
\midrule
GPT-5.5 & 63.5 & 70.4 & 71.3 & 53.3 & 58.0 \\
Gemini 3.5 Flash & 60.4 & 69.6 & 69.7 & 55.6 & 55.7 \\
Claude Opus 4.8 & 60.8 & 64.8 & 65.0 & 53.3 & 57.2 \\
\bottomrule
\end{tabularx}
}
  \caption{Per-category pass rates for the top three models.}
  \label{tab:competitor-cats}
  \end{table*}

Terminal-Bench-LILT enables analysis of model weaknesses by language and task category.
These comparisons are descriptive rather than controlled: empirical difficulty varies across languages, and category sizes are unequal.
Even with these limitations, the results reveal broad performance patterns worth investigating.

\paragraph{Per-Language Results}
Table~\ref{tab:competitor-langs} reports per-language pass rates for the top three models.
Performance varies substantially across languages, as do the model rankings.

Hindi and German are the two lowest-scoring languages for every model, both around 50--53\%.
The two leading models show different language strengths:
GPT-5.5 leads on Arabic, Spanish, Japanese, and Serbian, whereas Gemini~3.5~Flash leads on German and Korean.
The largest gaps between them occur in Japanese (13.4~points), Spanish (13.3~points), and Chinese (11.4~points).
Claude Opus~4.8 generally falls behind these two models but leads on Czech and Hindi.

Note that Gemini~3.5~Flash shows notably more uniform performance across languages than the others ($\sigma_{\mathrm{lang}}=4.9$), with no language below 52\%.
Among the ten languages, Hindi and Korean show the most consistent performance across models ($\sigma_{\mathrm{model}}=0.5$), whereas Japanese, Spanish, and Chinese show the largest cross-model variation.

\paragraph{Per-Category Results}
We also report pass rates for the top models by task category (Table~\ref{tab:competitor-cats}).
Text handling is the hardest category for every model, whereas text conversion and environment interaction are the easiest.
The gap between a model's best and worst category can be wide, reaching 18.0~points for GPT-5.5 between text conversion and text handling.

We hypothesize that this gap reflects how much language-specific knowledge each category requires.
For example, text conversion relies on standardized transformations with direct library support (e.g., Unicode normalization), so a general solution often transfers across languages.
In contrast, text handling depends on locale-specific rules that default tools handle incorrectly, such as Turkish case folding or stemming for agglutinative languages, forcing the model to supply the exact knowledge these tasks withhold.

\subsection{Ablation Studies}
\label{sec:ablation}

\paragraph{Effect of Instruction Language}
Every Terminal-Bench-LILT task is accompanied by an English translation of its native instruction, which lets us run the same task with the translated instruction while holding the data, environment, and verifier fixed.
This isolates the effect of switching the instruction language, comparing the target language against the models' strongest language, English.

\begin{figure}[!t]
\centering
\includegraphics[width=\columnwidth]{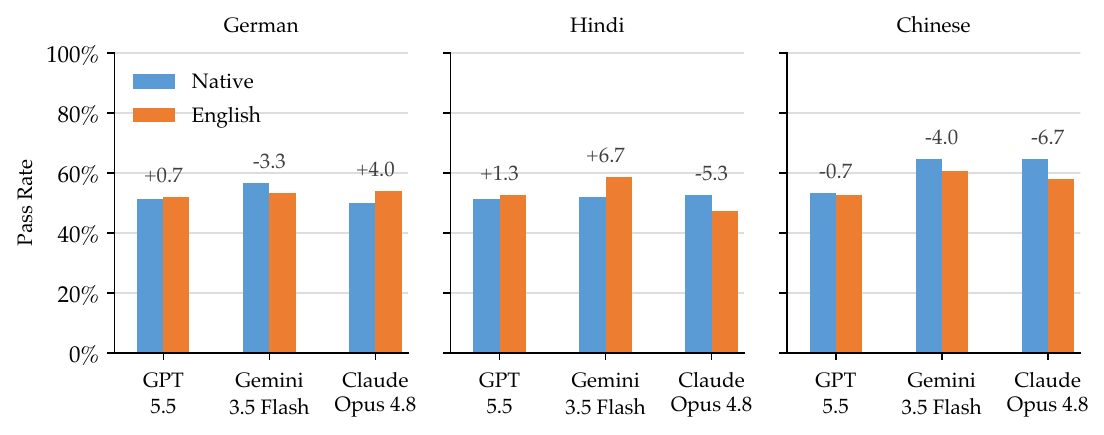}
\caption{Native language vs.\ English instruction ablation. Every bar pair covers the same 30~tasks, scored in both conditions. The number above each pair is the difference in pass rate, English minus native.}
\label{fig:instruction-lang}
\end{figure}

Figure~\ref{fig:instruction-lang} shows this comparison in three languages with different scripts, using the top three models from the main results (Section~\ref{sec:main}).
Across the nine model-language configurations, English instructions never move the pass rate by more than 7~points, and move it by less than 5 in six of them, similar to the small, inconsistent comprehension effects reported for translated benchmarks (Section~\ref{sec:related}).
Even with English instructions, every configuration stays around 50--60\%.
This confirms the premise behind our native-first task design: instruction language is not the bottleneck; the hard part is working with native-script data under locale-specific rules.

\paragraph{Effect of Reasoning}
\label{sec:reasoning}

\begin{figure}[!t]
\centering
\includegraphics[width=\columnwidth]{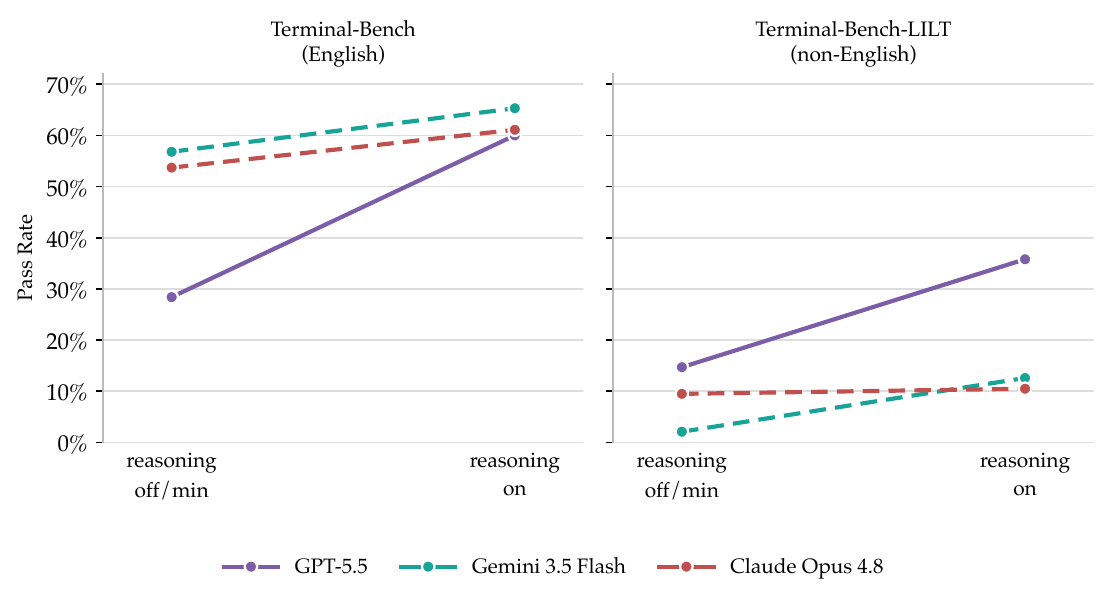}
\caption{Reasoning ablation. Solid lines mark significant changes with $p<.05$ in a two-sided paired permutation test.}
\label{fig:reasoning}
\end{figure}

All LLMs tested in this work are capable of reasoning via chain-of-thought \citep{wei2022chain}, which may improve problem solving on coding tasks at the cost of more output tokens and longer runtimes.
To study this effect, we ran the evaluation with reasoning switched on and off and compared the results.\footnote{When on, we used each provider's default reasoning level: \emph{medium} for GPT-5.5 and Gemini~3.5~Flash, \emph{high} for Claude Opus~4.8. Gemini~3.5~Flash cannot disable reasoning entirely by design, so for the off condition we set it to its \emph{minimal} level.}
For this ablation, we effectively removed the tasks' time limit, letting each agent run until it finished organically.
This study was run on 19 Terminal-Bench-LILT tasks that are hard under the empirical difficulty classification (Table~\ref{tab:dataset-stats}).
For comparison, we also ran the same reasoning ablation on 19 randomly sampled hard tasks from the original, English-only Terminal-Bench \citep{merrill2026terminalbench}; note that their tasks pose pure engineering challenges, unlike ours, which also require linguistic and cultural knowledge.

Figure~\ref{fig:reasoning} shows the results.
While reasoning improves GPT-5.5, no significant effect can be observed with Gemini~3.5~Flash or Claude Opus~4.8.
With reasoning on, the three models converge to a similar level on the English tasks (60.0--65.3\%), whereas on Terminal-Bench-LILT they range from 10.5\% to 35.8\%.
Reasoning thus lifts English performance to a shared level, but leaves large differences in multilingual capability across models.
Across models, reasoning inflates output token counts to 1.4--2.7$\times$ their off/minimal-reasoning levels, yet the large majority of the Terminal-Bench-LILT trials still fail.
Note that some tasks in German and Hindi stay at a 0\% pass rate even with reasoning.

\section{Conclusion}
\label{sec:conc}

We introduce Terminal-Bench-LILT, an agentic coding benchmark of 300 tasks written by native-speaker programmers in ten non-English languages.
Each task requires language-, region-, and culture-specific knowledge that a native developer takes for granted, built into everyday engineering work rather than tested in isolation.
A layered review pipeline retains only tasks that are realistic, verifiable, and genuinely locale-specific.
The benchmark is far from saturated: frontier models still fail over a third of trials, many tasks go unsolved entirely, pass rates vary widely by language, and rankings do not follow general coding-benchmark standing.
Our ablation studies show that this difficulty is not an artifact of non-English prompting or of limited compute.
Terminal-Bench-LILT reveals gaps in multilingual capability across models, languages, and task categories that English benchmarks cannot expose.
Its native-first design also offers a template for evaluation beyond English.

\section*{Ethics Statement}

First, all tasks were written by native-speaker programmers who contributed voluntarily with informed consent and compensation; we screened submissions for personal, proprietary, or sensitive data and collected no identifying information about contributors.
Second, our ten languages are a small, uneven sample and any single language spans many locales we do not fully cover, so low scores signal a measurement gap rather than a verdict on any language or its speakers.
Third, this is a diagnostic benchmark for a narrow capability: strong scores do not certify a model as fit for deployment, and weak scores should prompt further study, not exclusion of a language.
Fourth, we embed canary strings to limit training-data contamination and plan to refresh the pool over time, and because these tasks execute code, all evaluation runs in sandboxed containers.

\bibliographystyle{colm2026_conference}
\bibliography{references}

\appendix

\section{Sample Tasks}
\label{sec:appendix-task-list}

\renewcommand{\arraystretch}{1.25}
\setlength{\tabcolsep}{4pt}
\fontsize{8.5}{10}\selectfont
\captionsetup{font=normalsize}
\begin{longtable}{>{\raggedright\arraybackslash}p{0.24\textwidth} >{\raggedright\arraybackslash}p{0.16\textwidth} >{\raggedright\arraybackslash}p{0.413\textwidth} >{\raggedright\arraybackslash}p{0.10\textwidth}}
\caption{Sample tasks in the benchmark.}
\label{tab:appendix-task-list} \\
\toprule
\textbf{Task ID} & \textbf{Category} & \textbf{Description} & \textbf{Difficulty} \\
\midrule
\endfirsthead

\multicolumn{4}{c}%
{{\tablename\ \thetable{} -- continued from previous page}} \\
\toprule
\textbf{Task ID} & \textbf{Category} & \textbf{Description} & \textbf{Difficulty} \\
\midrule
\endhead

\midrule
\multicolumn{4}{r}{{Continued on next page}} \\
\endfoot

\bottomrule
\endlastfoot

% inlined from data/task_list.tex by build_arxiv.py
\texttt{cs-\allowbreak implicit-\allowbreak alphabet-\allowbreak frequency} & Cultural \& Other & Count occurrences of each Czech alphabet letter in a text without being told what the Czech alphabet is — the agent must implicitly know all 42 letters including accented characters and the `ch' digraph. & Medium \\
\texttt{cs-\allowbreak nbsp-\allowbreak justify} & Cultural \& Other & Split Czech sentences into xelatex-justified lines with no Overfull hbox, while honoring Czech orthotypographic non-breaking-space and soft-hyphen rules. & Hard \\
\texttt{cs-\allowbreak rozuctovani-\allowbreak tepla} & Cultural \& Other & Fix a script that audits a Czech SVJ heating-cost allocation against vyhlaska 269\textasciislash{}2015 Sb. & Medium \\
\texttt{de-\allowbreak bgb-\allowbreak citation-\allowbreak resolver} & Cultural \& Other & Resolve a list of German civil-code citations (e.g. § 433 Abs. 1 Satz 2) against a BGB XML file. Parse the hierarchical legal-text format, count Sätze per Absatz while respecting period-terminated legal abbreviations (Abs., Nr., S., gem., i.V.m.), and return each citation's exact text as JSON. & Medium \\
\texttt{de-\allowbreak umsatzsteuer-\allowbreak klassifikation} & Cultural \& Other & Classify German products by statutory VAT rate (7\% reduced or 19\% standard) under §12 UStG and Anlage 2 (Stand 1.1.2026), fixing a broken keyword draft. Many names are German compounds where the Grundwort sets the rate while the Bestimmungswort misleads, so decompose the compound first. & Hard \\
\texttt{es-\allowbreak locale-\allowbreak search-\allowbreak analyzer} & Text Handling & Build a Spanish search analyzer that indexes a corpus, answers queries with locale-correct collation, and emits snippets that preserve the original surface form. & Easy \\
\texttt{ja-\allowbreak terminal-\allowbreak width} & Inter\-na\-tio\-nal\-iza\-tion & Wrap each line of a NOC-log file to a fixed 40-column width: strip ANSI escapes, fold halfwidth katakana to fullwidth (merging {\cjkfont ﾞﾟ}), measure width (CJK\textasciislash{}fullwidth\textasciislash{}emoji = 2, ASCII = 1, variation-selector clusters = one 2-wide unit), never splitting a wide char. Bytes stay verbatim, so no NFC\textasciislash{}NFKC. & Medium \\
\texttt{sr-\allowbreak cadastral-\allowbreak parcel-\allowbreak ledger} & Text Conversion & Normalize Serbian cadastral parcel request text across registry aliases, script variants, area units, ownership shares, and Serbian collation into a JSON ledger. & Easy \\
\texttt{sr-\allowbreak locale-\allowbreak sort} & Cultural \& Other & Fix a buggy Serbian Latin sorter so the digraphs dž, lj and nj are treated as single letters in any word position, not only at word start. & Easy \\
\texttt{tr-\allowbreak username-\allowbreak dedup-\allowbreak casefold} & Text Handling & Detect duplicate user registrations in a Turkish CRM by applying the Turkish-locale-aware username canonicalization (NFC + I-pair pre-fold + casefold) that Python's str.lower() and str.casefold() do not implement by default. & Hard \\
\end{longtable}
\normalsize

\end{document}